# Comparative Analysis of Document-Level Embedding Methods for Similarity Scoring on Shakespeare Sonnets and Taylor Swift Lyrics

Klara Krämer

**Abstract:** This study evaluates the performance of TF-IDF weighting, averaged Word2Vec embeddings, and BERT embeddings for document similarity scoring across two contrasting textual domains. By analysing cosine similarity scores, the methods' strengths and limitations are highlighted. The findings underscore TF-IDF's reliance on lexical overlap and Word2Vec's superior semantic generalisation, particularly in cross-domain comparisons. BERT demonstrates lower performance in challenging domains, likely due to insufficient domain-specific fine-tuning.

## 1 Introduction

Document similarity assessment plays an important role in various natural language processing (NLP) applications, such as information retrieval, plagiarism detection, recommendation systems, and question answering [11, 19]. For instance, in recommendation systems, document similarity helps personalise suggestions by finding content that closely matches user preference. These tasks rely on accurate measurements of how similar documents are in terms of their structure, content, and meaning, which depends on the way the document is represented computationally. This representation is usually done in vector format and is obtained via document embedding methods. Various methodologies can be employed to obtain document-level embeddings, and the choice of method directly impacts the accuracy and usefulness of the similarity scores calculated [14, 19].

This study compares three methods of document-level embeddings for document similarity scoring: Averaged Word2Vec [18], TF-IDF weighting [16], and BERT embeddings [6]. Each method offers distinct advantages and limitations, and this comparative study applies these techniques to a diverse set of texts - specifically, the sonnets of William Shakespeare and the lyrics of Taylor Swift. Generating document-level embeddings for documents in these two contrasting genres and analysing their similarity scores will help evaluate how effective, reliable, robust, and adaptable different embedding techniques are for scoring document similarity.

The methodology employed in this project involves creating document-level cosine similarity matrices for each dataset and method. This is followed by a quantitative analysis, in which the average similarity scores within each matrix are calculated and compared.

# 2 Background and Related Work

There is a large body of research on document similarity measures, with different studies focusing on various aspects of document similarity and different applications.

Traditionally, methods such as TF-IDF (Term Frequency-Inverse Document Frequency) as proposed by Sparck Jones have been employed to compute the similarity between documents [16]. While TF-IDF captures relative term relevance within documents, the method cannot encode word meaning and contextual information, limiting its effectiveness for nuanced language like poetry or lyrics [13].

The more recent advancement Word2Vec, developed by Mikolov et al., maps words to continuous vector spaces based on their co-occurring context words [18]. This approach captures semantic relationships between words, but how to use these word-level embeddings to create whole-document embeddings remains an open research question. Proposed solutions include the Paragraph Vector model Doc2Vec [9], and simple but effective methods such as averaging word embeddings to obtain sentence- and document-level embeddings from their constituent words [20]. Addressing these limitations, Devlin et al. introduced BERT (Bidirectional Encoder Representation from Transformers), a pre-trained model that generates context-sensitive embeddings [6].

Most studies on document similarity scoring focus on specific application domains, for instance, patent-to-patent comparisons. Younge et al. showed that a TF-IDF-based vector space model outperforms traditional patent classifiers in terms of accuracy, specificity, and generality [21]. However, they did not compare this approach with other vectorisation models.

In a study building on the above, Shahmirzadi et al. analysed the performance of different vector space models in measuring semantic text similarity on patents [14]. Comparing TF-IDF weighting, topic modelling, and Doc2Vec embeddings, they found that pre-trained and highly tuned neural embeddings like Doc2Vec provided the best results. However, they also observed a trade-off between computational efficiency and accuracy, noting that a simple TF-IDF approach was more practical for their application. The authors of this paper noted a lack of studies evaluating the performance of different vectorisation methods on more challenging datasets, which this paper aims to address.

Within the domain of Shakespeare sonnets and Pop lyrics, there is only limited NLP research. Some existing papers utilised clustering approaches to attempt to answer the Shakespeare authorship question [1] [1, 2]. Koppel and Seidman applied first- and second-order document similarity measures to Shakespeare plays in a similar authorship verification task [8]. Their evaluation used 40 plays attributed to Shakespeare, leaving room for the future work this paper focuses on, as a larger corpus of shorter documents such as sonnets may yield different results for document similarity measures.

In a study on music similarity, Knees and Schedl noted that lyrics-based features can sometimes outperform advanced measures such as audio-based features in assessing song semantics and similar-ity [7]. A more domain-specific paper examined linguistic patterns in Taylor Swift's writing, focusing on word and 3-gram co-occurrences. They conducted a quantitative study on these lyrics based on the assumption that song lyrics often originate as lyric poems [5], an assumption that this paper will help evaluate by calculating similarity scores between Shakespearean poetry and Swift lyrics.

[1] The Shakespeare authorship question asks whether all works commonly attributed to William Shakespeare have truly been written by him, or whether there have been ghostwriters or false attributions.

# 3 Research Question

As the above body of related work indicates, document similarity scoring is an interesting area of research which so far is quite unexplored within specialised domains such as poetry and lyrics. Therefore, this paper investigates the effectiveness of different techniques in generating document similarity scores for these domains.

**Research Question:** How do the three embedding methods under evaluation - namely averaged Word2Vec, TF-IDF weighting, and BERT embeddings - affect the document similarity scores generated within and between these domains?

To establish this, the methods are to be applied to (1) a dataset of Taylor Swift lyrics, (2) a dataset of Shakespeare sonnets, and (3) a combination of the above two in one dataset.

Specifically, the following hypotheses are to be tested:

1. TF-IDF weighting will show higher average similarity scores within the Taylor Swift dataset than the other two datasets, reflecting the straightforward narrative style of the lyrics.

2. Averaging the Word2Vec embeddings of the words within each document will result in higher similarity scores for Taylor Swift songs than for Shakespeare sonnets. This is likely because of the more contemporary language of the Swift lyrics, which aligns better with the inherent Word2Vec training data.

3. Averaged Word2Vec embeddings will assign higher scores to Swift-Shakespeare pairs in the joint dataset than the TF-IDF method since Word2Vec can recognise similarities in meaning even when the vocabulary differs significantly.

4. BERT embeddings will perform better than the other methods at identifying similarities between the Shakespeare sonnets due to BERT's ability to capture the contextual nuances in poetic language.

These hypotheses will be evaluated by calculating document-level cosine similarity matrices using each method on each of the 3 datasets. These matrices will then be analysed quantitatively by calculating and comparing their average similarity scores. To test hypothesis 3, the average cosine similarity of Swift-Shakespeare pairs in the joint dataset will be compared across the different methods. In addition, qualitative error analysis will be performed in cases in which the results are unexpected.

The datasets on which the above analyses will be performed are taken from GitHub, utilising the repositories by Marquez [10] and Finch [4] for the Shakespeare sonnets and Swift lyrics respectively.

Since the Swift dataset contains 232 songs, but there are only 154 sonnets by William Shakespeare, 154 of the Swift songs are selected at random to ensure equal numbers across the two datasets. See Table 8 in the Appendix for a list of the Taylor Swift songs that were excluded. Some minor exploratory data analysis (EDA) was performed to get an overview of the data loaded from the two sources.

The results of this EDA can be seen in Table 1.

| **Dataset** | **Shakespeare** | **Swift** |
|---|---|---|
| Total number of words | 17507 | 56932 |
| Vocabulary size | 3027 | 3734 |
| Lexical diversity[2] | 0.1729 | 0.0656 |
| Top content words[3] | 'thy', 'thou', 'love', 'thee', 'doth', 'beauty', 'time', 'mine', 'shall', 'heart' | 'like', 'oh', 'know', 'never', 'time', 'one', 'love', 'could', 'back', 'got' |

**Table 1:** EDA on the two core datasets

# 4 Experimental Design

Based on the above general Research Question of how different embedding methods impact document similarity scores, and the concrete hypotheses stated, the experiments to test these hypotheses can be designed.

All document embedding methods will be applied to the same input data. To prepare for the experiments, all sonnets and songs will be preprocessed through lowercasing, tokenisation, and the removal of stopwords and punctuation. Two derived datasets will be created: The 'Combined Dataset' including all sonnets and songs, and the 'Distinct Dataset', which contains only sonnet-song similarity scores. The following procedure will be applied to each of the three embedding methods under evaluation:

1. Create embeddings of all documents for each dataset. The specific parameters used for each embedding method can be seen in the Appendix in Table 3. Tokens for which no embedding exists will be treated as zero-vectors.

2. Compute cosine similarity scores[4] for all document pairs for each dataset, storing the results in similarity matrices.

3. Construct a similarity matrix for the 'Distinct Dataset', where rows correspond to Shakespeare sonnets and columns correspond to Swift songs. This matrix does thus not contain similarity scores between individual sonnets or between songs, but only between sonnet-song pairs. Table 4 in the Appendix visualises this difference by showing the layout of the similarity matrices generated from these two datasets.

[2]Lexical diversity = (Vocabulary size) / (Total number of words)
[3]Excluding common stopwords as defined by *nltk*
[4]To compute the similarity scores, the *sklearn.metrics.pairwise.cosine_similarity* function is used

Given this experimental design, the embedding method and dataset type are independent variables, since they determine how document embeddings are generated and compared. The similarity scores and the means derived from them are the measurable outcomes influenced by the choice of embedding method and dataset, making them the dependent variables.

# 5 Results

This section will be separated into four subsections, each corresponding to one of the four hypotheses. Table 2 gives an overview of the mean similarity scores for each embedding method and dataset. Notably, TF-IDF yields significantly lower similarity scores in all cases, while Word2Vec produces the highest scores.

| | Shakespeare | Taylor Swift | Combined Dataset | Distinct Dataset |
|---|---|---|---|---|
| **Averaged Word2Vec** | 0.9539 | 0.9993 | 0.9998 | 0.9997 |
| **TF-IDF** | 0.1157 | 0.1486 | 0.1075 | 0.0734 |
| **BERT** | 0.9207 | 0.9088 | 0.7983 | 0.6820 |

**Table 2:** Mean Similarity Scores for each Embedding Method and Dataset

## 5.1 TF-IDF Similarity Scores

The first null hypothesis, which states that the mean similarity scores derived using TF-IDF weighting are identical across the Shakespeare, Swift, and Combined datasets, is rejected based on the results of a one-way ANOVA performed on the three similarity matrices. The p-value for this analysis is less than 0.001, providing sufficient evidence to reject the null hypothesis and support the alternative hypothesis. Applying Tukey's honestly significant difference (HSD) test confirms that the average similarity scores within the Taylor Swift dataset are significantly higher than those in the other two datasets, as shown in Figure 1.

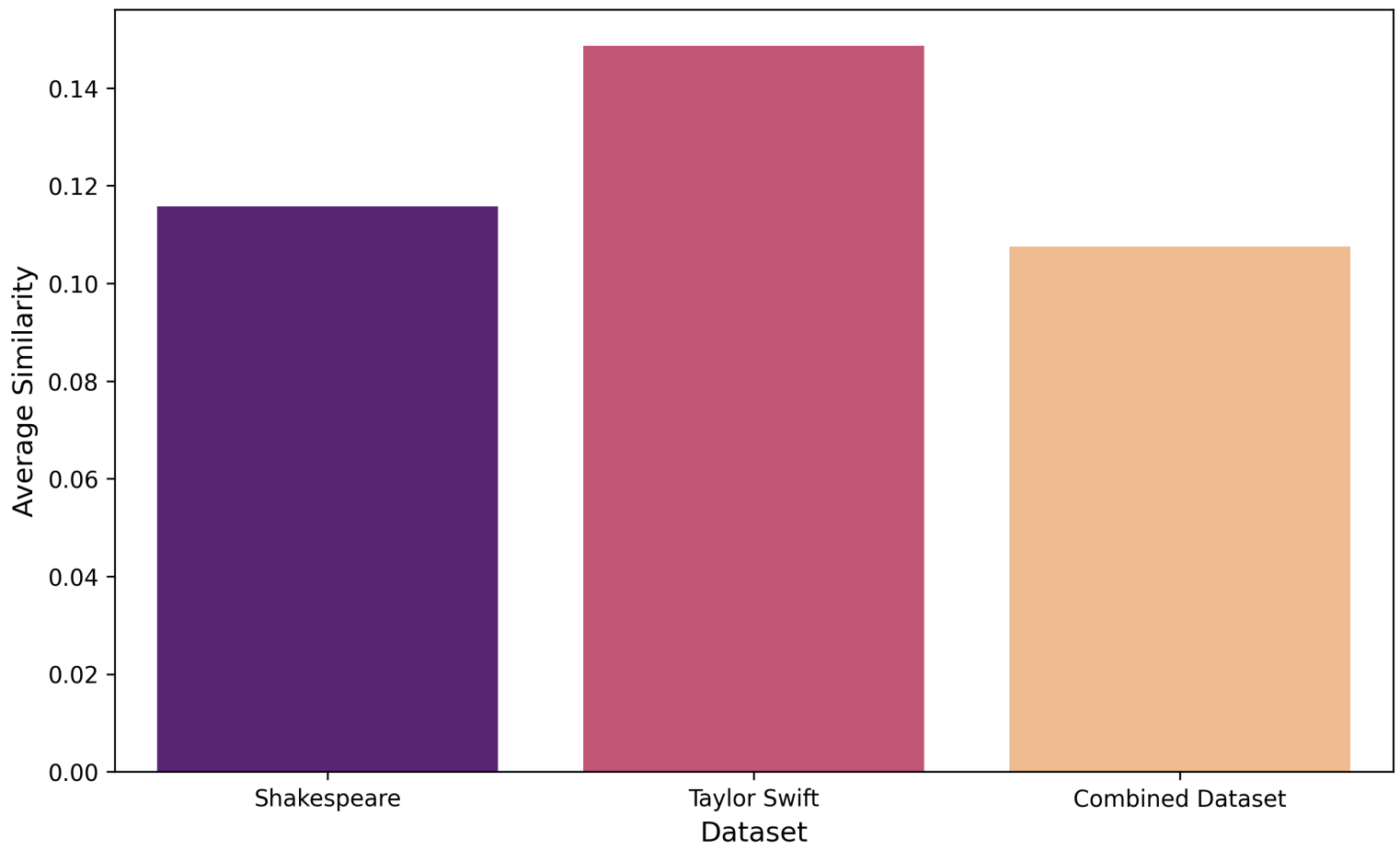

**Figure 1:** Average similarity scores produced using TF-IDF

## 5.2 Word2Vec Similarity Scores

The second null hypothesis is that the mean similarity scores produced using Word2Vec are equivalent for the Shakespeare and Swift datasets. To evaluate this, a Mann-Whitney test is conducted. The resulting p-value is substantially greater than 0.05 ($p \approx 1$), indicating no statistical evidence to reject the null hypothesis. This suggests comparable scores across both datasets, contrary to the hypothesis that averaged Word2Vec would yield higher similarity scores for Taylor Swift songs. Figure 2 highlights the minimal variance in these similarity scores.

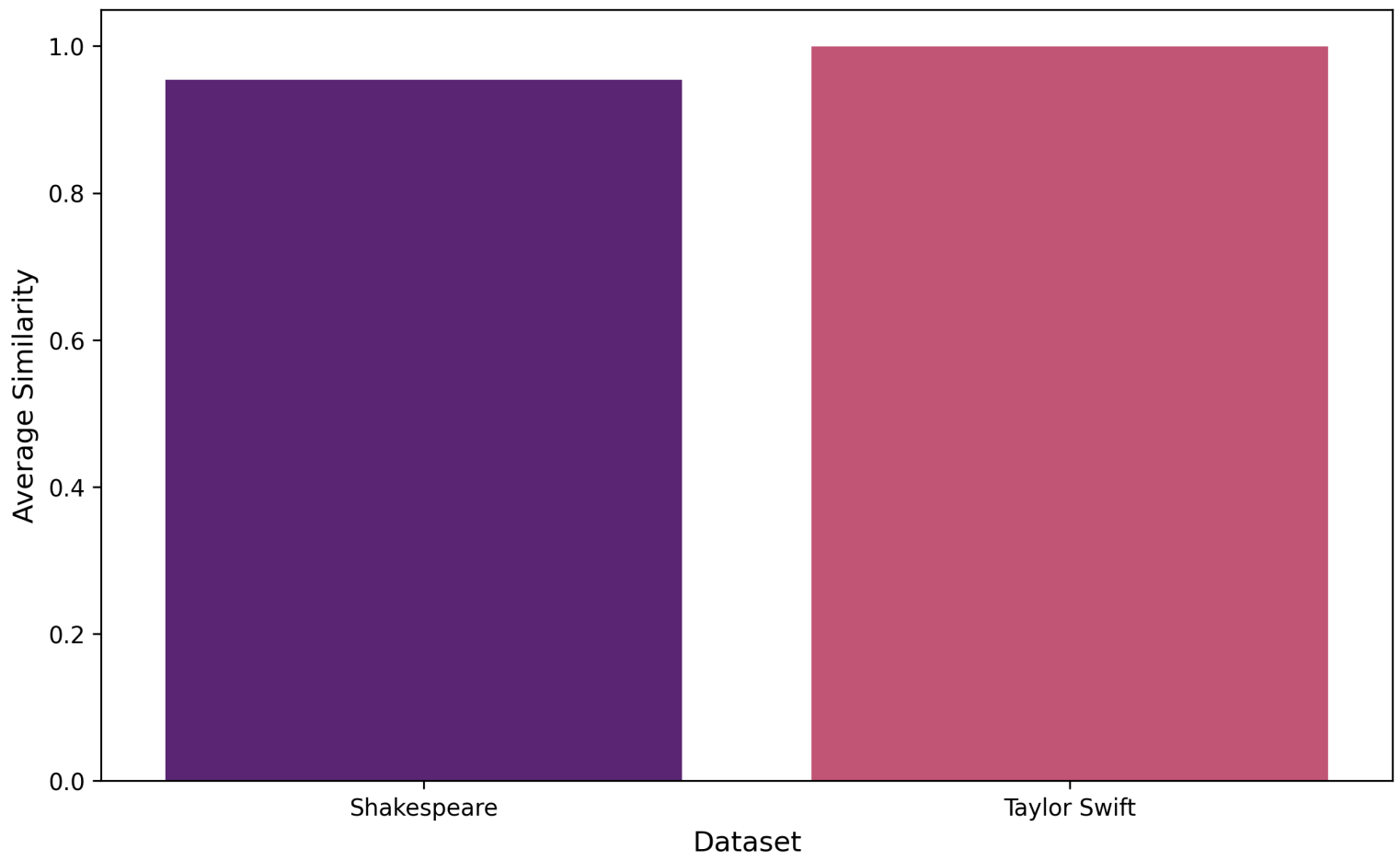

**Figure 2:** Average similarity scores produced using Word2Vec

## 5.3 Similarity Scores of Shakespeare-Swift Pairs

The third null hypothesis, namely that the mean similarity scores within the Distinct dataset are the same for averaged Word2Vec embeddings and TF-IDF, is tested via a Wilcoxon test. The test yields a p-value of less than 0.001, meaning that the null hypothesis is rejected. As can be seen in Figure 3, the Word2Vec method yields significantly higher similarity scores than the TF-IDF method, supporting the hypothesis that Word2Vec effectively captures semantic similarity across disparate vocabularies, which TF-IDF does not.

## 5.4 Similarity Scores in the Shakespeare Dataset

To test the final null hypothesis, which states that the mean similarity scores within the Shakespeare dataset are the same across methods, a one-way ANOVA is used, yielding $p < 0.001$.

However, Tukey's HSD test reveals that BERT does not produce the highest scores. This contradicts the hypothesis that BERT would identify more similarities in the Shakespeare sonnets compared to the other methods. Figure 4 presents the results.

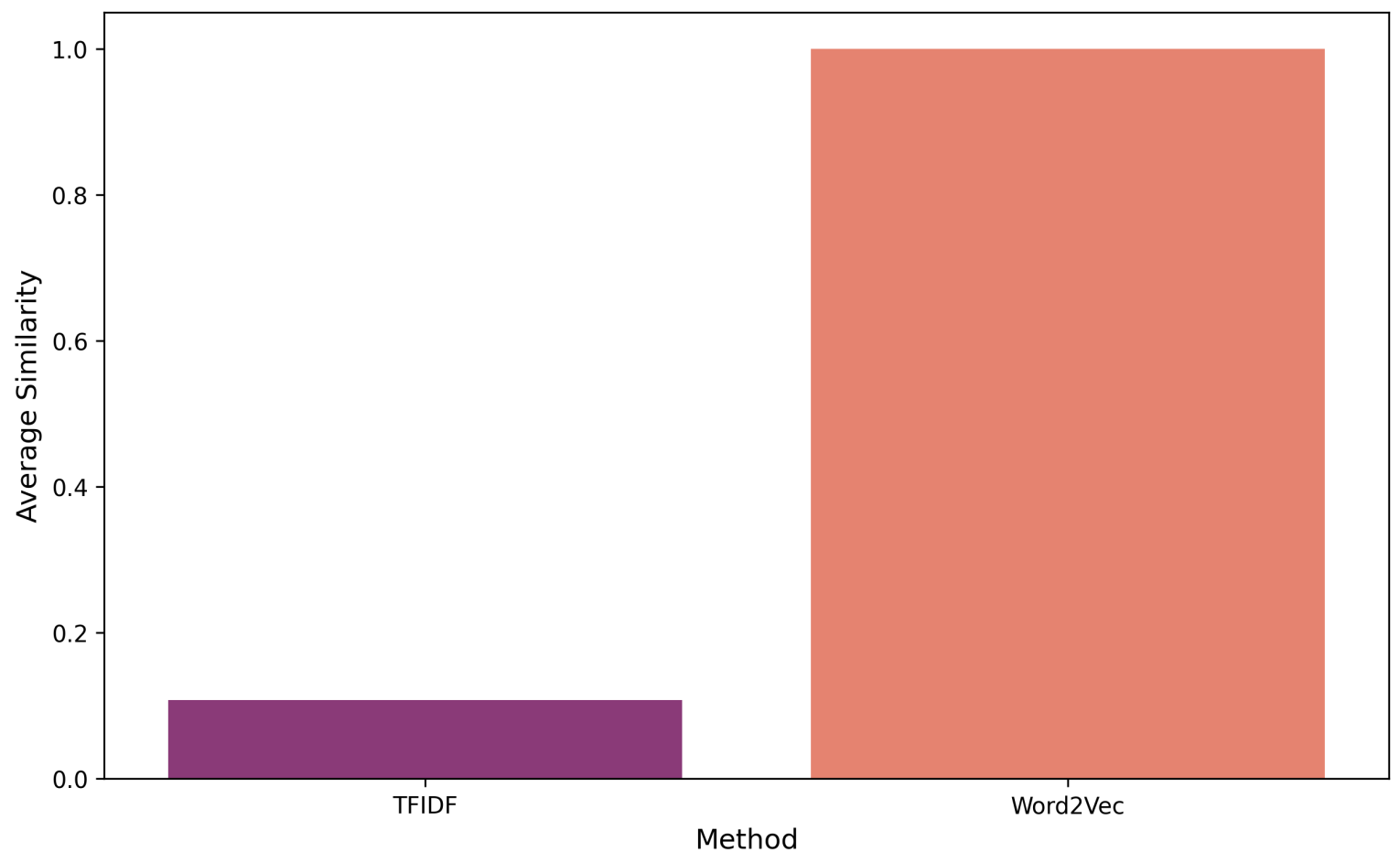

**Figure 3:** Average similarity scores of Shakespeare-Swift pairs in the Distinct dataset

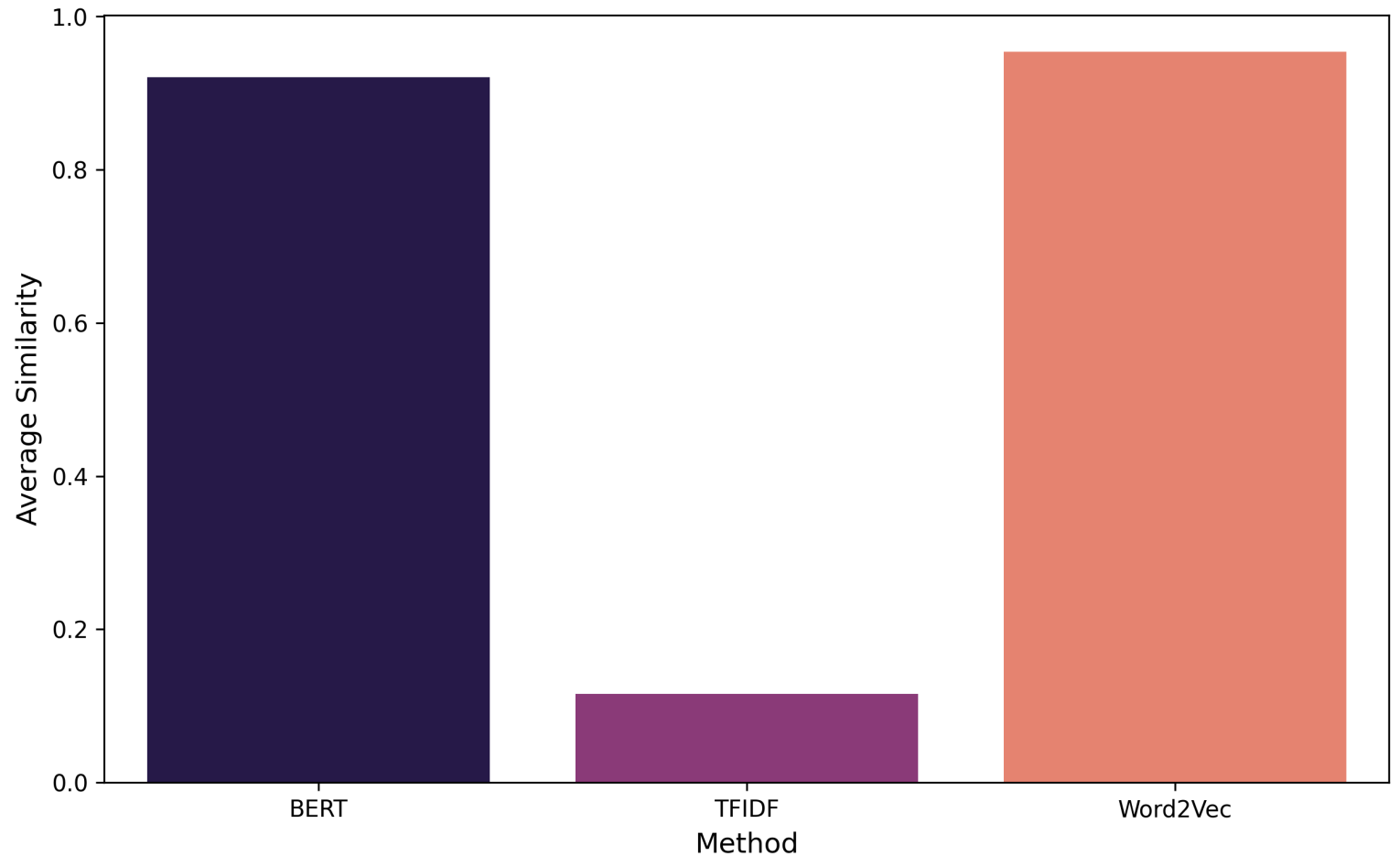

**Figure 4:** Average similarity scores within the Shakespeare dataset

# 6 Discussion and Future Work

As presented in the above Results section, this study provides insights into the effectiveness of different document-level embedding methods in generating document similarity scores across specialised domains. Each hypothesis was statistically tested, and the findings highlight both the capabilities and limitations of the embedding methods in capturing relationships within and across datasets.

TF-IDF demonstrates strong performance in identifying lexical overlap in straightforward texts such as Taylor Swift lyrics, but is less effective with Shakespeare's nuanced and metaphorical language, highlighting its limited contextual understanding.

Word2Vec does not yield higher similarity scores for Swift songs than for Shakespeare sonnets, contrary to expectations. Future work might build upon this by examining Word2Vec's performance on texts from several centuries rather than just two.

Qualitative analysis provides additional insights. Word2Vec identified the highest similarity between Shakespeare's Sonnet 60 and Sonnet 149[5], both of which explore themes of unreciprocated love. The lowest similarity scores were assigned to Taylor Swift's "Out of the Woods" and "Shake It Off"[6], reflecting differences in tone and message, with the former song reflecting emotional struggles and the latter emphasising empowerment and joy. This outlier analysis provides insight into potential reasons for the comparatively high and low similarity scores, even though they deviate from the original hypothesis.

The results for the third hypothesis highlight the comparative strengths of Word2Vec over TF-IDF in recognising semantic similarities. The higher similarity scores for Shakespeare-Swift document pairs in the Distinct dataset confirm that Word2Vec effectively captures deeper semantic relations even when vocabulary differs.

BERT embeddings do not yield the highest similarity scores for the Shakespeare dataset, contradicting the hypothesis that BERT's contextualised representations would prove most successful at capturing the intricate nature of Shakespeare sonnets. This suggests limitations in BERT's performance for poetic texts, potentially due to insufficient domain-specific fine-tuning or inadequate representation of such language in its pre-training. Future research could address this by evaluating BERT on larger, more diverse poetic datasets and exploring different fine-tuning approaches.

An analysis of the most dissimilar pair of Shakespeare sonnets defined by BERT, namely Sonnet 97 and Sonnet 148[7] shows shared themes of devotion and idolisation but differing tones: The former emphasises melancholic longing and idealised beauty, while the latter critiques love's irrationality with an accusatory tone.

The study is not without its limitations. The above results point to correlations between the methods' performance and the characteristics of the datasets, though causation cannot be fully established. The observed performance differences may be influenced not only by the embedding techniques themselves but also by factors such as pre-training data, hyperparameter selection, and dataset size. Additionally, the Shakespeare dataset contains fewer documents than the Taylor Swift corpus, and while random subsampling was used to balance the datasets, this approach may have introduced variability in the results. The exclusive use of cosine similarity as a metric may not fully capture the nuances of semantic similarity, especially for complex poetic language. A more in-depth qualitative evaluation of document similarity could provide a richer understanding of how the different embedding methods perform on these specialised domains.

Future work could also explore hybrid approaches that combine lexical and semantic features to build on the above results. Another interesting area of future research is performing similar evaluations using a wider range of different embedding methods such as FastText [3] and SenteceBERT [12].

[5]The sonnets are given in the Appendix in Table 5

[6]The lyrics are given in the Appendix in Table 7

[7]See Table 6 in the Appendix for the sonnets

# 7 Conclusion

This study explored the effectiveness of TF-IDF, averaged Word2Vec, and BERT embeddings for document similarity scoring in the domains of Taylor Swift lyrics and Shakespeare sonnets to evaluate the embedding methods' strengths and weaknesses.

The results showed that TF-IDF was most effective when applied to Swift lyrics, while it struggled with the metaphorical language of Shakespeare sonnets. Averaged Word2Vec embeddings performed better at capturing semantic similarity, particularly for Swift-Shakespeare pairs, but did not show significant differences between the Shakespeare and Swift datasets. Contrary to expectations, BERT did not identify more similarities within the Shakespeare dataset than the other methods.

There is a range of future work that could build on this research to develop a deeper and more nuanced understanding of how different embedding methods perform on diverse datasets within the context of similarity scoring.

# 8 Appendix

| Embedding Method | Parameters |
|---|---|
| Averaged Word2Vec | vector_size=100, window=5, min_count=1, workers=4 |
| TFIDF | standard vectoriser.fit_transform() usage |
| BERT | from_pretrained('bert-base-uncased'); return_tensors='pt', truncation=True, padding=True, max_length=512 |

**Table 3:** Parameters used for each embedding method

| Dataset Name | Similarity Matrix Layout |
|---|---|
| Combined Dataset | 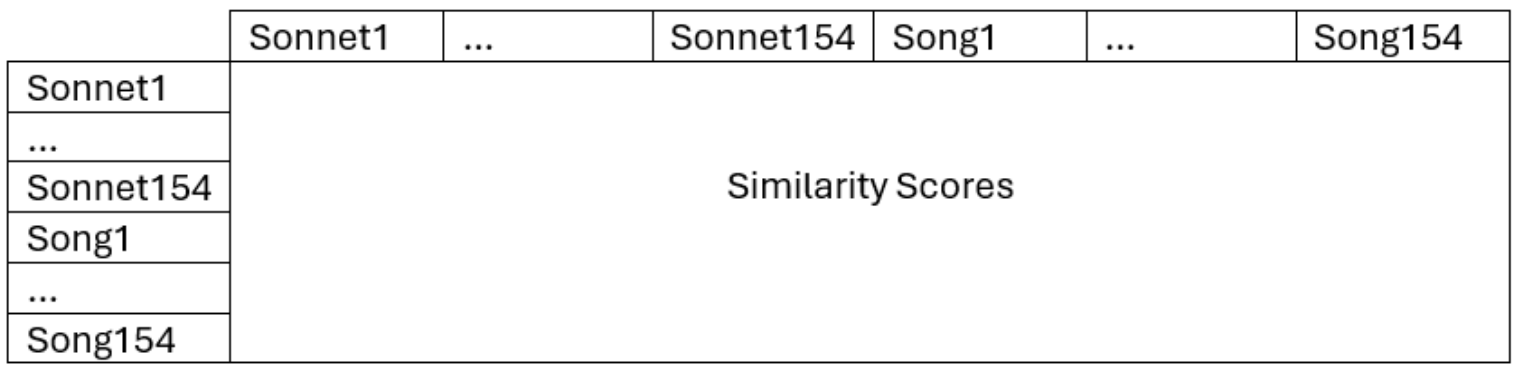 |
| Distinct Dataset | 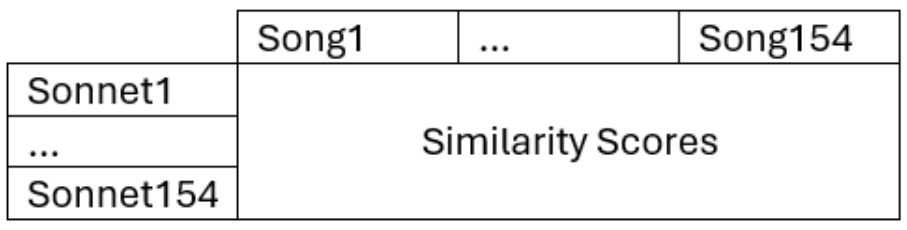 |

**Table 4:** Similarity matrix layouts for the Combined and Distinct datasets

| Sonnet | Text |
|---|---|
| Sonnet 60 [15] | Is it thy will, thy image should keep open<br>My heavy eyelids to the weary night?<br>Dost thou desire my slumbers should be broken,<br>While shadows like to thee do mock my sight?<br>Is it thy spirit that thou send'st from thee<br>So far from home into my deeds to pry,<br>To find out shames and idle hours in me,<br>The scope and tenor of thy jealousy?<br>O, no! thy love, though much, is not so great:<br>It is my love that keeps mine eye awake:<br>Mine own true love that doth my rest defeat,<br>To play the watchman ever for thy sake:<br>For thee watch I, whilst thou dost wake elsewhere,<br>From me far off, with others all too near. |
| Sonnet 149 [15] | O! from what power hast thou this powerful might,<br>With insufficiency my heart to sway?<br>To make me give the lie to my true sight,<br>And swear that brightness doth not grace the day?<br>Whence hast thou this becoming of things ill,<br>That in the very refuse of thy deeds<br>There is such strength and warrantise of skill,<br>That, in my mind, thy worst all best exceeds?<br>Who taught thee how to make me love thee more,<br>The more I hear and see just cause of hate?<br>O! though I love what others do abhor,<br>With others thou shouldst not abhor my state:<br>If thy unworthiness raised love in me,<br>More worthy I to be beloved of thee. |

**Table 5:** Selected Shakespeare sonnets for outlier analysis for hypothesis 2

| Sonnet | Text |
|---|---|
| Sonnet 97 [15] | From you have I been absent in the spring,<br>When proud pied April, dressed in all his trim,<br>Hath put a spirit of youth in every thing,<br>That heavy Saturn laughed and leapt with him.<br>Yet nor the lays of birds, nor the sweet smell<br>Of different flowers in odour and in hue,<br>Could make me any summer's story tell,<br>Or from their proud lap pluck them where they grew:<br>Nor did I wonder at the lily's white,<br>Nor praise the deep vermilion in the rose;<br>They were but sweet, but figures of delight,<br>Drawn after you, you pattern of all those.<br>Yet seemed it winter still, and you away,<br>As with your shadow I with these did play. |
| Sonnet 148 [15] | Canst thou, O cruel! say I love thee not,<br>When I against myself with thee partake?<br>Do I not think on thee, when I forgot<br>Am of my self, all tyrant, for thy sake?<br>Who hateth thee that I do call my friend,<br>On whom frown'st thou that I do fawn upon,<br>Nay, if thou lour'st on me, do I not spend<br>Revenge upon myself with present moan?<br>What merit do I in my self respect,<br>That is so proud thy service to despise,<br>When all my best doth worship thy defect,<br>Commanded by the motion of thine eyes?<br>But, love, hate on, for now I know thy mind,<br>Those that can see thou lov'st, and I am blind. |

**Table 6:** Selected Shakespeare sonnets for outlier analysis for hypothesis 4

| Song | Lyrics |
|---|---|
| Out Of The Woods [17] | Looking at it now, it all seems so simple<br>We were lying on your couch, I remember<br>You took a Polaroid of us, then discovered (Then discovered)<br>The rest of the world was black and white, but we were in screaming color<br>[Chorus:]<br>And I remember thinking<br>Are we out of the woods yet?, are we out of the woods yet?, are we out of the woods yet?, are we out of the woods?<br>Are we in the clear yet?, are we in the clear yet?, are we in the clear yet?, in the clear yet, good<br>Are we out of the woods yet?, are we out of the woods yet?, are we out of the woods yet?, are we out of the woods?<br>Are we in the clear yet?, are we in the clear yet?, are we in the clear yet?, in the clear yet, good<br>Looking at it now, last December (Last December)<br>We were built to fall apart, then fall back together (Back together)<br>Oh, your necklace hanging from my neck, the night we couldn't quite forget when we decided, we decided<br>To move the furniture so we could dance<br>Baby, like we stood a chance<br>Two paper airplanes flying, flying, flying<br>[Chorus]<br>Remember when you hit the brakes too soon?<br>Twenty stitches in a hospital room<br>When you started crying, baby I did too<br>But when the sun came up I was looking at you<br>Remember when we couldn't take the heat?<br>I walked out, I said, "I'm setting you free"<br>But the monsters turned out to be just trees<br>When the sun came up you were looking at me<br>You were looking at me, oh<br>You were looking at me<br>[Chorus]<br>[Chorus] |
| Shake It Off [17] | I stay out too late<br>Got nothing in my brain<br>That's what people say, mm-mm<br>That's what people say, mm-mm<br>I go on too many dates<br>But I can't make 'em stay<br>At least that's what people say, mm-mm<br>That's what people say, mm-mm<br>[Chorus:]<br>But I keep cruisin'<br>Can't stop, won't stop movin'<br>It's like I got this music in my mind<br>Sayin' it's gonna be alright<br>'Cause the players gonna play, play, play, play, play<br>And the haters gonna hate, hate, hate, hate, hate<br>Baby, I'm just gonna shake, shake, shake, shake, shake<br>I shake it off, I shake it off (hoo-hoo-hoo)<br>Heartbreakers gonna break, break, break, break, break<br>And the fakers gonna fake, fake, fake, fake, fake<br>Baby, I'm just gonna shake, shake, shake, shake, shake<br>I shake it off, I shake it off (hoo-hoo-hoo)<br>I never miss a beat<br>I'm lightnin' on my feet<br>And that's what they don't see, mm-mm<br>That's what they don't see, mm-mm<br>I'm dancin' on my own (dancin' on my own)<br>I make the moves up as I go (moves up as I go)<br>And that's what they don't know, mm-mm<br>That's what they don't know, mm-mm<br>[Chorus]<br>Hey, hey, hey<br>Just think, while you've been gettin' down and out about the liars<br>And the dirty, dirty cheats of the world<br>You could've been gettin' down to this sick beat<br>My ex-man brought his new girlfriend<br>She's like, "Oh my God!" but I'm just gonna shake<br>And to the fella over there with the hella good hair<br>Won't you come on over, baby? We can shake, shake, shake (yeah)<br>[Chorus] |

**Table 7:** Selected Swift songs for outlier analysis

| Album | Title |
|---|---|
| Taylor Swift | Tim McGraw |
| Taylor Swift | Teardrops On My Guitar |
| Taylor Swift | A Place In This World |
| Taylor Swift | Cold As You |
| Taylor Swift | The Outside |
| Taylor Swift | Mary’s Song (Oh My My My) |
| Taylor Swift | Our Song |
| Taylor Swift | Invisible |
| Taylor Swift | A Perfectly Good Heart |
| Fearless (Taylor’s Version) | Fifteen |
| Fearless (Taylor’s Version) | Tell Me Why |
| Fearless (Taylor’s Version) | Jump Then Fall |
| Fearless (Taylor’s Version) | Come In With The Rain |
| Fearless (Taylor’s Version) | Superstar |
| Fearless (Taylor’s Version) | That’s When |
| Fearless (Taylor’s Version) | Bye Bye Baby |
| Speak Now (Taylor’s Version) | Speak Now |
| Speak Now (Taylor’s Version) | Mean |
| Speak Now (Taylor’s Version) | The Story Of Us |
| Speak Now (Taylor’s Version) | Enchanted |
| Speak Now (Taylor’s Version) | Long Live |
| Speak Now (Taylor’s Version) | Timeless |
| Red (Taylor’s Version) | State Of Grace |
| Red (Taylor’s Version) | I Knew You Were Trouble |
| Red (Taylor’s Version) | I Almost Do |
| Red (Taylor’s Version) | The Lucky One |
| Red (Taylor’s Version) | Everything Has Changed |
| Red (Taylor’s Version) | Come Back...Be Here |
| Red (Taylor’s Version) | Girl At Home |
| Red (Taylor’s Version) | Ronan |
| Red (Taylor’s Version) | Nothing New |
| Red (Taylor’s Version) | I Bet You Think About Me |
| Red (Taylor’s Version) | Forever Winter |
| 1989 (Taylor’s Version) | Welcome To New York |
| 1989 (Taylor's Version) | This Love |
| 1989 (Taylor's Version) | You Are In Love |
| 1989 (Taylor's Version) | "Slut!" |
| 1989 (Taylor's Version) | Say Don't Go |
| 1989 (Taylor's Version) | Is It Over Now? |
| 1989 (Taylor's Version) | Sweeter Than Fiction |
| Reputation | ...Ready For It? |
| Reputation | End Game |
| Reputation | I Did Something Bad |
| Reputation | Delicate |
| Reputation | Getaway Car |
| Reputation | Dancing With Our Hands Tied |
| Reputation | This Is Why We Can't Have Nice Things |
| Lover | I Forgot That You Existed |
| Lover | Paper Rings |
| Lover | Death By A Thousand Cuts |
| Lover | Afterglow |
| Lover | It's Nice To Have A Friend |
| Folklore | The 1 |
| Folklore | Cardigan |
| Folklore | The Last Great American Dynasty |
| Folklore | This Is Me Trying |
| Folklore | Illicit Affairs |
| Folklore | Mad Woman |
| Folklore | Hoax |
| Evermore | Dorothea |
| Evermore | Cowboy Like Me |
| Evermore | Long Story Short |
| Midnights (The Til Dawn Edition) | Midnight Rain |
| Midnights (The Til Dawn Edition) | Bejeweled |
| Midnights (The Til Dawn Edition) | Labyrinth |
| Midnights (The Til Dawn Edition) | Karma |
| Midnights (The Til Dawn Edition) | Paris |
| Midnights (The Til Dawn Edition) | Would've, Could've, Should've |
| The Tortured Poets Department (The Anthology) | My Boy Only Breaks His Favorite Toys |
| The Tortured Poets Department (The Anthology) | Fresh Out The Slammer |
| The Tortured Poets Department (The Anthology) | Florida!!! |
| The Tortured Poets Department (The Anthology) | Guilty As Sin? |
| The Tortured Poets Department (The Anthology) | Clara Bow |
| The Tortured Poets Department (The Anthology) | The Albatross |
| The Tortured Poets Department (The Anthology) | Chloe Or Sam Or Sophia Or Marcus |
| The Tortured Poets Department (The Anthology) | So High School |
| The Tortured Poets Department (The Anthology) | The Prophecy |
| The Tortured Poets Department (The Anthology) | Cassandra |

**Table 8:** Swift Songs that were excluded from the dataset